%% file: main.tex
\documentclass[10pt,twocolumn,letterpaper]{article}

\usepackage[pagenumbers]{cvpr}     %

\definecolor{cvprblue}{rgb}{0.21,0.49,0.74}
\usepackage[pagebackref,breaklinks,colorlinks,allcolors=cvprblue]{hyperref}
\usepackage{array}

\title{Rethinking Planning Recovery in Quantized Latent World Models}

\author{Chunzheng Li\\
Soochow University\\
\and
Jin Wang\\
Soochow University\\
\and
Siao Liu\\
Soochow University\\
\and
Zesheng Jia\\
Soochow University\\
\and
Daqi Liu\\
AGIBOT\\
\and
Dongchun Ren\\
Yootta \\
\and
Jianping Wang\\
City University of Hong Kong\\
}

\begin{document}
\maketitle
\input{sec/0_abstract}

\input{sec/1_intro}

\input{sec/2_Related_work}

\input{sec/3_Prob_setup}
\input{sec/4_recover}

\input{sec/5_analysis}

\input{sec/6_Conclusion}

{
    \small
    \bibliographystyle{ieeenat_fullname}
    \bibliography{main}
}

\clearpage
\input{sec/X_suppl}


\end{document}

%% file: sec/0_abstract.tex
\begin{abstract}
Low-bit weight quantization reduces the storage and transfer costs of latent world models for planning (LWM-P), but can substantially degrade their closed-loop planning performance. Yet the extent to which this degradation can be recovered across different world models remains unclear. Moreover, closed-loop success rates alone offer limited insight into the sources of planning degradation, motivating more controlled analysis of quantization-induced distortions. We first investigate recoverability through post-training calibration, showing that combining local reconstruction with global full-precision (FP) supervision can restore 4-bit weight-quantized (W4) planning performance to near-FP levels across diverse models and tasks. However, both the initial degradation and the effectiveness of recovery strategies vary substantially across architectures. To investigate these differences, we model measured quantization residuals and perform controlled distortion injection at different strengths and interfaces, together with cross-model transfer. Our controlled interventions reveal that mean squared prediction error alone cannot fully characterize planning sensitivity, as error location and distribution also influence closed-loop success. Cross-model distortion transfer further reveals that models with stronger W4 performance can be more sensitive to controlled perturbations, while those with poorer W4 performance can remain resilient, highlighting the distinct roles of distortion generation and planning tolerance.
\end{abstract}

%% file: sec/1_intro.tex
\section{Introduction}
\label{sec:introduction}

Latent world models for planning (LWM-P) enable visual planning by encoding observations and predicting the effects of candidate actions in representation space~\cite{zhou2024dinowm,maes2026leworldmodel,zhang2026hjepa,zhou2026idwm,zhou2026injepa}.
A planner evaluates candidate action sequences through latent rollouts and selects actions according to their predicted outcomes.
As these evaluations recur across search iterations and control steps, model storage, computation, and memory traffic become important constraints on practical deployment.
Post-training quantization (PTQ) compresses pretrained models using limited calibration data, without full model retraining~\cite{nagel2020adaround,wei2022qdrop,yuan2022ptq4vit,frantar2023gptq,li2023repqvit}. Weight quantization reduces model storage and transfer requirements and can improve inference efficiency with appropriate execution kernels~\cite{lin2024awq}.
However, quantization introduces numerical errors that can degrade planning performance.
Recovering this performance is therefore essential for realizing the efficiency benefits of quantization in LWM-P.
\begin{figure}[t]
    \centering
    \includegraphics[width=\columnwidth]{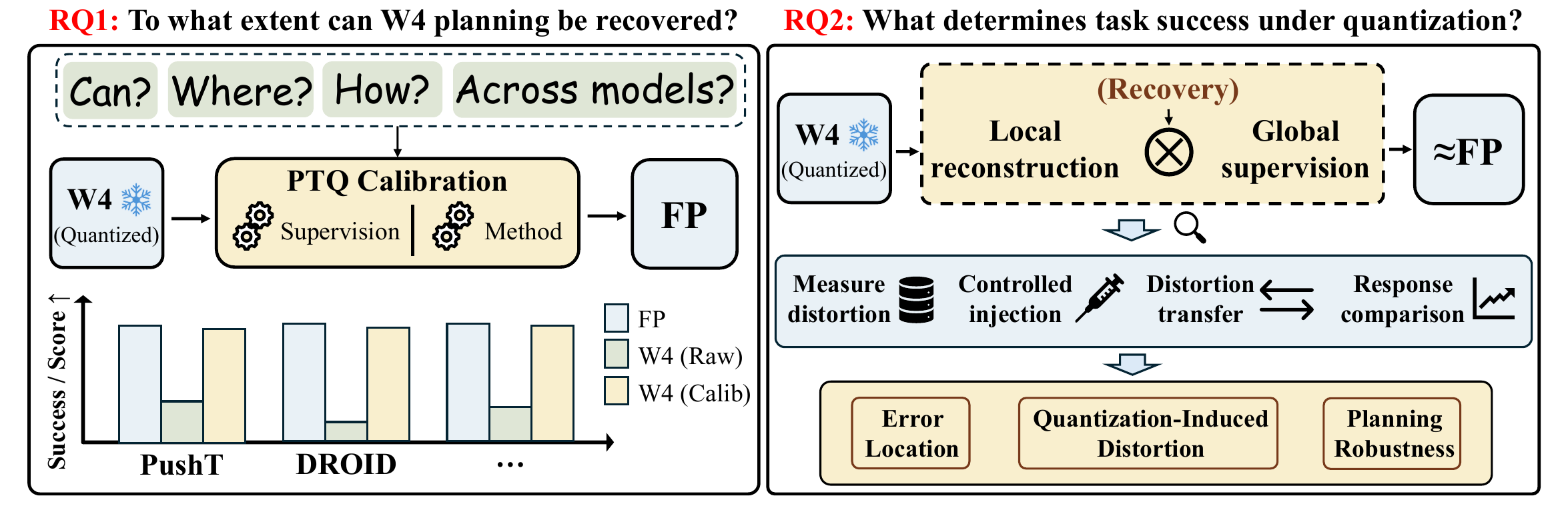}
    \vspace{-15pt}
    \caption{Overview of planning recovery and analysis in quantized LWM-P. Left: local reconstruction and global FP supervision recover W4 planning performance across models and tasks. Right: controlled distortion analysis examines how error location, distortion profiles, and planning sensitivity affect task success.}
    \label{fig:recovery_overview}
    \vspace{-18pt}
\end{figure}
QuantWM~\cite{fu2026quantwm} investigates quantization granularity and asymmetric encoder--predictor sensitivity in DINO-WM, while mixed-bit studies examine precision allocation and planner search budgets~\cite{ranganath2026wherebits}. However, the extent to which planning performance can be recovered under 4-bit weight quantization (W4) across diverse LWM-P architectures remains insufficiently understood. This motivates our first research question \textbf{RQ1: To what extent can W4 planning performance be recovered?}

To answer RQ1, we evaluate W4 planning recovery across LWM-Ps with different representation and dynamics designs. Our experiments show that local reconstruction combined with global full-precision (FP) supervision can restore W4 planning performance to near-FP levels across diverse models and tasks. However, the initial quantization degradation and the effectiveness of calibration vary substantially across architectures, raising questions about the factors governing planning performance under quantization.

Closed-loop task success measures the outcome of quantization but offers limited insight into the sources of performance degradation. A model may retain strong W4 performance because quantization introduces relatively small latent distortions, because its planning system tolerates those distortions, or both. Conversely, severe W4 degradation does not necessarily imply greater sensitivity to controlled perturbations. These ambiguities motivate our second research question \textbf{RQ2: What determines task success under quantization?} We investigate this question from three perspectives: \textbf{where} quantization errors arise, \textbf{what} distortions they introduce into latent representations, and \textbf{how} the planning system responds to these distortions (\Cref{fig:recovery_overview}).

Our systematic empirical study combines W4 planning recovery experiments across diverse architectures with controlled latent-space distortion analysis. To answer RQ2, we model measured quantization residuals and inject them into fixed world models at controlled strengths and interface locations, allowing us to carefully examine planning responses to specific distortions. We further transfer residuals across models to distinguish the distortions generated by quantization from the receiving models' sensitivity to them. Our experiments show that mean squared prediction error alone cannot fully characterize planning sensitivity, as error location and distribution also affect closed-loop task success. More strikingly, cross-model transfer reveals that models with stronger W4 performance can be more sensitive to controlled perturbations than models suffering greater degradation under direct quantization.

Our contributions are twofold:
\begin{itemize}
    \item \textbf{Experimental disentanglement of distortion generation and planning sensitivity.} Through controlled residual injection and cross-model transfer, we experimentally separate quantization-induced distortions from the planning system's response to them. Our results reveal that prediction error magnitude alone cannot fully characterize planning sensitivity, and that stronger W4 performance does not necessarily imply greater tolerance to controlled perturbations.
    \item \textbf{Planning recovery across world models.} We systematically investigate W4 planning recovery across diverse LWM-P architectures and tasks, demonstrating that local reconstruction combined with global FP supervision can restore near-FP planning performance while revealing model-dependent degradation and recovery behaviors.
\end{itemize}

%% file: sec/2_Related_work.tex
\section{Related Work}
\label{sec:related_work}

\paragraph{Latent world models for planning.}
Latent world models enable planning by evaluating imagined action
sequences in a learned representation space, as established by
PlaNet~\cite{hafner2019planet} and extended to goal-conditioned
visual planning with pretrained features by
DINO-WM~\cite{zhou2024dinowm}.
Recent LWM-Ps adopt diverse representation-learning strategies.
C-JEPA learns interaction-dependent dynamics through object-level
latent masking~\cite{nam2026cjepa}.
PLDM jointly learns compact representations and latent dynamics
with variance--covariance regularization~\cite{sobal2025learning},
while LeWM employs end-to-end predictive learning with isotropic
Gaussian regularization derived from
LeJEPA~\cite{maes2026leworldmodel,balestriero2025lejepa}.
Efficiency-oriented architectures further modify latent dynamics:
DDP-WM separates primary dynamics from
background updates~\cite{yin2026ddpwm},
while Scope-WM focuses on prediction-relevant
tokens and promising action
sequences~\cite{li2026scopewm}.
Both provide distinct architectures for
studying quantization recovery.

\paragraph{Post-training quantization and calibration.}
Post-training quantization (PTQ) compresses pretrained networks
using limited calibration data.
Mean squared error (MSE)-based quantization selects scaling parameters to reduce tensor approximation error~\cite{choukroun2019lowbit},
while AdaRound learns weight-rounding
decisions~\cite{nagel2020adaround} and BRECQ reconstructs
network blocks to account for interlayer
dependencies~\cite{li2021brecq}.
AWQ protects activation-salient weights~\cite{lin2024awq}, SmoothQuant smooths activation outliers by shifting quantization difficulty to weights through equivalent transformations~\cite{xiao2023smoothquant}, and OmniQuant learns clipping thresholds and equivalent transformations~\cite{shao2024omniquant}.
Beyond local reconstruction, PD-Quant uses full-model prediction
differences to guide quantization-parameter
optimization~\cite{liu2023pdquant}.
Building on these calibration principles, we compare round-to-nearest (RTN), optimized mean squared error (OMSE), AWQ, SmoothQuant, OmniQuant, and blockwise reconstruction with 4-bit weights and 32-bit
activations (W4A32), and examine how local reconstruction
and global FP supervision contribute to planning recovery
across diverse LWM-P architectures.

\paragraph{Quantization of world models and embodied systems.}
QuantWM provides an empirical investigation of weight-only and
weight--activation PTQ on DINO-WM, revealing the importance
of quantization granularity, asymmetric encoder--predictor
sensitivity, and planning-objective
alignment~\cite{fu2026quantwm}.
A subsequent mixed-bit study further examines precision
allocation and planning budgets~\cite{ranganath2026wherebits}.
Related embodied-model studies investigate low-bit action
generation.
QuantVLA calibrates attention and output scales in
vision-language-action models~\cite{zhang2026quantvla}.
QuantWAMs incorporates model structure, rollout distributions,
and task objectives into world-action model
calibration~\cite{zhou2026quantwams}, while Q-WAM protects
action-sensitive subspaces using a higher-precision
low-rank branch~\cite{akbari2026qwam}.
These studies establish the importance of task-aware quantization, motivating evaluation of planning recovery across diverse LWM-Ps.

\paragraph{Quantization distortions and planning robustness.}
Prediction fidelity and downstream decision quality need not
coincide, as studied in decision-focused learning and
ranking-based formulations~\cite{mandi2022decisionrank}.
For latent world models, AD-WM investigates action discrimination and decision quality under counterfactual predictions~\cite{qiu2026adwm}, while decision-metric alignment studies examine whether latent goal distances preserve candidate-action rankings~\cite{wang2026decisionmetric}.
These findings motivate examining how
representation errors affect downstream planning.
We study weight quantization by modeling measured
latent residuals, injecting them at controlled
strengths into fixed world models, and
transferring them across architectures.
The resulting responses characterize the
effects of error location, distortion profiles,
and planning robustness on task success.

%% file: sec/3_Prob_setup.tex
\section{Preliminaries and Experimental Setup}
\label{sec:preliminaries}
This section introduces the common framework of LWM-P, the quantization and recovery settings, and the models, tasks, and evaluation protocols used in our experiments.
\subsection{Latent World Models for Planning}

\begin{figure*}[t]
    \centering
    \includegraphics[width=\textwidth]{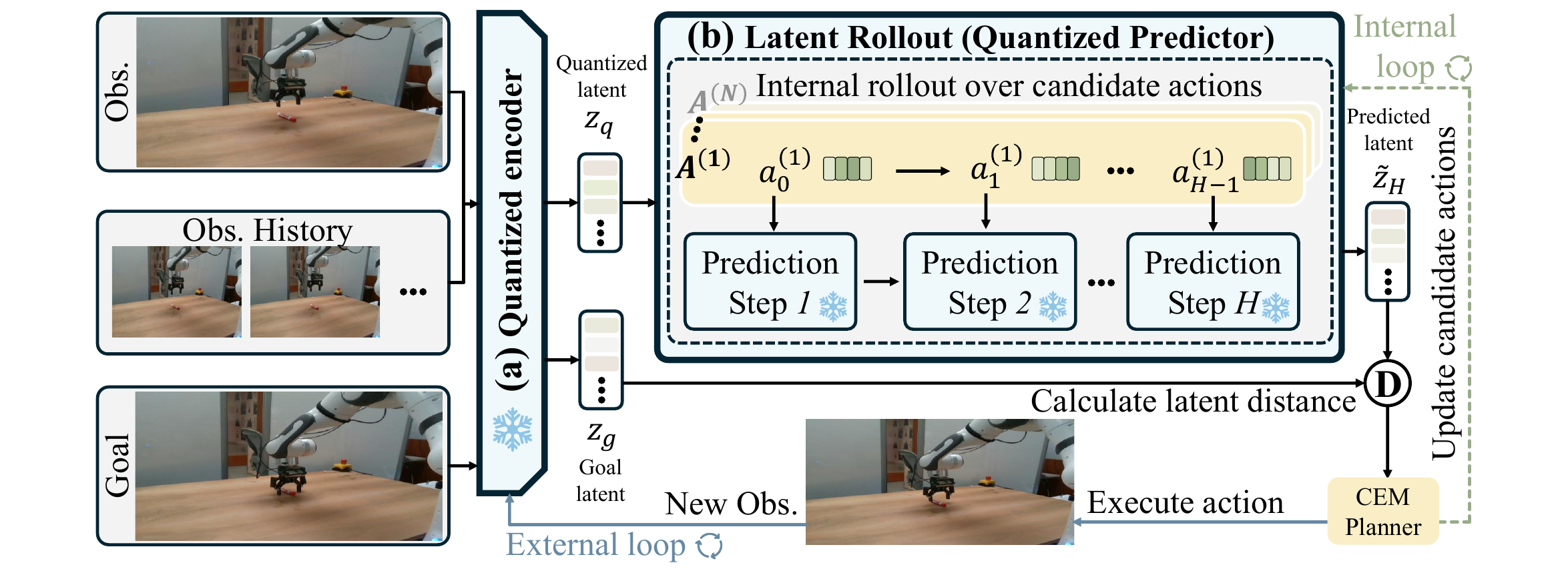}
    \vspace{-20pt}
    \caption{Planning with quantized LWM-P. (a) The encoder maps observation histories and goals into latent representations. (b) The predictor rolls out candidate action sequences, and the planner scores their predicted outcomes against the goal. The inner CEM loop updates the candidate-action distribution, while the outer MPC loop executes an action block, receives new observations, and replans.}
    \label{fig:planning_pipeline}
    \vspace{-10pt}
\end{figure*}

Despite differences in representation learning and dynamics modeling, the LWM-Ps studied here share the general planning pipeline illustrated in \Cref{fig:planning_pipeline}, consisting of an encoder, an action-conditioned predictor, and a planner. At control step $t$, the encoder $E_{\theta}$ maps the current observation and its history $O_t=(o_{t-K+1},\ldots,o_t)$ into latent states $Z_t=E_{\theta}(O_t)$, and encodes the goal observation $o_g$ as $z_g=E_{\theta}(o_g)$. The predictor $P_{\phi}$ then imagines future latent states under candidate action sequences. The exact latent structure and observation history remain model-dependent.

Specifically, a cross-entropy method (CEM) planner
samples $N$ candidate action sequences
$\mathbf{A}_t^{(i)}=(a_t^{(i)},\ldots,a_{t+H-1}^{(i)})$
over a planning horizon $H$.
For each candidate, the predictor recursively
generates the corresponding future latent states:
\begin{equation}
    \hat z_{t+h+1}^{(i)}
    =P_{\phi}\!\left(
        \mathcal H_{t+h}^{(i)},a_{t+h}^{(i)}
    \right),
    \quad h=0,\ldots,H-1,
    \label{eq:latent_rollout}
\end{equation}
where $\mathcal H_{t+h}^{(i)}$ denotes the latent history available at prediction step $h$, including previously predicted states.
Each candidate is evaluated by a model-specific
goal-conditioned cost, written schematically as
$J_t^{(i)}=\mathcal C(\hat z_{t+H}^{(i)},z_g)$.
The cost may incorporate spatial masks, proprioceptive
terms, or other model-specific scoring mechanisms.
Within the inner planning loop, CEM selects the lowest-cost candidates as an elite set and updates the mean and variance of its sampling distribution over multiple iterations.
The planner selects an action sequence from the
optimized candidate distribution according to its
model-specific CEM implementation.
In the outer model-predictive control (MPC) loop, the first action block of this sequence is executed in the environment, after which new observations are obtained for the next planning cycle.
Thus, encoder errors affect the initial state and goal representations, while predictor errors can accumulate across imagined steps and alter candidate comparisons and subsequent planning decisions.

\subsection{Quantization Policy and Recovery Setup}
\label{sec:quantization_recovery}

We primarily investigate weight-only W4A32 PTQ, with additional precision settings considered in auxiliary comparisons. We use group-wise quantization, where G64 denotes a group size of 64 weights, and retain model-specific quantization granularity where applicable. For a weight group $W_g$, a representative symmetric $b$-bit quantizer is
\begin{equation}
    Q_b(W_g;s_g)=s_g\,\mathrm{clip}\!\left(
        \mathrm{round}(W_g/s_g),
        -2^{b-1}+1,\,2^{b-1}-1\right),
    \label{eq:weight_quantization}
\end{equation}
where $s_g$ is the group scale; calibrated variants may also optimize weight-rounding decisions.
Quantization is applied to eligible encoder and predictor weights according to each model's coverage policy, while activations and unquantized components retain floating-point precision.

We compare FP, round-to-nearest (RTN) W4, and calibrated W4 configurations. Our main recovery experiments examine local reconstruction of quantized blocks and global FP supervision of encoded states and action-conditioned multi-step predictions. Calibration updates quantizer scales and rounding parameters while keeping the pretrained FP reference and planner fixed. We also compare PTQ baselines and interface compensation, which applies learned corrections to encoder or predictor latent outputs. We evaluate planning performance primarily through closed-loop
task success, complemented by FP-referenced multi-step
prediction errors and candidate-cost agreement on shared
action pools for each model.

\subsection{Models, Tasks, and Evaluation Protocol}
\label{sec:evaluation_setup}

\textbf{Models and tasks.} Our principal evaluation includes C-JEPA, DINO-WM, LeWM, DDP-WM, and Scope-WM, covering different representation-learning and latent-dynamics designs. This selection allows us to compare quantization degradation and recovery across architectures under a shared planning task. We additionally evaluate PLDM on a smaller cohort to extend the comparison to another end-to-end trained world model.
We select Push-T as our principal closed-loop benchmark because its contact-rich manipulation requires coordinating object localization, contact placement, pushing direction, and multi-step displacement. Public checkpoints and evaluation code are available across the studied LWM-Ps, enabling comparisons of diverse architectures on a shared task. Each model retains its native observation processing, goal construction, planner cost, and stopping criterion.
For DROID, we evaluate CEM action planning on recorded trajectories using an offline Action Score that compares predicted action displacements with recorded actions, rather than physical execution success. Additional tasks are included when their corresponding checkpoints, FP baselines, and evaluation protocols are available.

\noindent\textbf{Closed-loop protocol.} For the controlled comparisons, we hold the action horizon, CEM population, elite count, optimization budget, MPC execution budget, and success criterion fixed across FP and W4 configurations within each model and task. We retain model-specific planner settings and use matched initial states and goals, with shared planner seeds where supported. For ranking analysis, FP and quantized models evaluate identical candidate actions to isolate differences in predicted costs and candidate selection.

\noindent\textbf{Evaluation size and reporting.} We use closed-loop success rate as the primary planning metric and evaluate the principal Push-T configurations on 300-episode cohorts. Smaller evaluations are reported separately with their corresponding cohort sizes.

For controlled distortion experiments, we fix the receiving model and planner while varying the strength, source, and injection interface of quantization-derived latent perturbations. We compare closed-loop success on matched tasks to characterize model-specific planning responses. Implementation details, hardware configurations, and evaluation settings are provided in the supplementary material.

%% file: sec/4_recover.tex
\section{Recovering Planning Performance in Quantized LWM-P}
\label{sec:recovery}

To address RQ1, we evaluate W4 planning recovery across different world models and examine the effects of local reconstruction and global FP supervision.

\subsection{Patterns of W4 Planning Recovery}
\label{sec:recovery_feasibility}

\begin{figure}[t]
    \centering
    \includegraphics[width=\columnwidth]{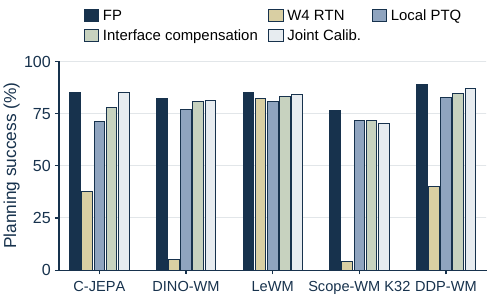}
    \vspace{-20pt}
    \caption{W4 planning recovery across five LWM-Ps on Push-T, comparing FP, W4 RTN, local PTQ, interface compensation, and joint calibration with global FP supervision.}
    \label{fig:cross_model_recovery}
    \vspace{-15pt}
\end{figure}

\Cref{fig:cross_model_recovery} compares FP, W4 round-to-nearest (RTN), and calibrated W4 planning success across five LWM-Ps: C-JEPA~\cite{nam2026cjepa}, DINO-WM~\cite{zhou2024dinowm}, LeWM~\cite{maes2026leworldmodel}, DDP-WM~\cite{yin2026ddpwm}, and Scope-WM~\cite{li2026scopewm}. Initial W4 degradation differs sharply across models. LeWM retains 82.3\% success under RTN, close to its FP baseline of 85.3\%, while C-JEPA and DDP-WM drop to 37.7\% and 40.3\%, respectively.
Despite similarly severe degradation, C-JEPA and DDP-WM show different recovery profiles: local reconstruction brings DDP-WM close to FP, whereas C-JEPA improves from 71.3\% after local reconstruction to 85.0\% with global FP supervision.
Scope-WM with 32 selected latent tokens (K32) exhibits another pronounced recovery, from 4.3\% under RTN to 71.7\% after G64 fitting.
For DINO-WM, interface compensation improves on local reconstruction.

\subsection{The Role of Calibration Supervision}
\label{sec:recovery_supervision}

To examine how FP supervision at different locations affects recovery, we compare local block reconstruction with global supervision of the assembled world model. Local reconstruction matches the outputs of quantized blocks to their FP references, optimizing grouped weight scales and rounding parameters~\cite{li2021brecq}. Global supervision instead aligns encoded observation and goal states and action-conditioned multi-step predictions with the FP model. Implementation details are provided in the supplementary material.

Global FP supervision substantially changes the relative performance of PTQ initializations (\Cref{tab:ptq_supervision}). OMSE rises from 17.0\% to 82.0\%, exceeding OmniQuant's 77.3\% despite OmniQuant's much stronger initial result of 71.3\%. G64 block MSE reaches 85.0\% from 71.3\%, matching FP.

More importantly, global FP supervision substantially reduces the performance disparity across PTQ initializations. Before global calibration, the six methods span 17.0--71.3\% success, a gap of 54.3 percentage points; afterward, the range narrows to 75.3--85.0\%, only 9.7 points. The ordering also changes: OMSE initially trails OmniQuant by 54.3 points but surpasses it by 4.7 points after calibration. These results show that local reconstruction quality alone does not determine final planning performance.

\begin{table}[t]
\centering
\caption{C-JEPA W4A32 planning success under different PTQ methods. Joint calibration combines PTQ with global FP supervision. The FP32 model size is 105.68 MB.}
\vspace{-8pt}
\label{tab:ptq_supervision}
\setlength{\tabcolsep}{3pt}
\begin{tabular}{@{}cccc@{}}
\toprule
Method & PTQ (\%) & Joint Calib. (\%) & Size (MB) \\
\midrule
RTN
    & 37.7 & 81.0 & 17.13 \\
OMSE~\cite{choukroun2019lowbit}
    & 17.0 & 82.0 & 17.13 \\
AWQ~\cite{lin2024awq}
    & 33.0 & 75.3 & 17.41 \\
SmoothQuant~\cite{xiao2023smoothquant}
    & 30.0 & 78.7 & 17.41 \\
OmniQuant~\cite{shao2024omniquant}
    & 71.3 & 77.3 & 17.13 \\
G64 block MSE
    & \textbf{71.3} & \textbf{85.0} & \textbf{17.13} \\
\bottomrule
\vspace{-20pt}
\end{tabular}
\end{table}

\Cref{tab:supervision_ablation} examines which global FP targets contribute most to C-JEPA recovery. Removing state alignment lowers success from 85.0\% to 78.0\%, and removing future visual prediction lowers it to 77.3\%. Removing future proprioceptive supervision has a much smaller effect, yielding 84.7\% success.
The ablation results also reveal a clear asymmetry among supervision targets. State alignment constrains the representations of current observations and goals, which provide the reference for evaluating candidate trajectories. Future visual supervision instead constrains action-conditioned predictions used to assess candidate outcomes. Removing either produces a substantial performance loss, whereas removing future proprioceptive supervision has little effect. This pattern identifies state representation and visual dynamics prediction as the two principal supervision targets associated with successful W4 planning recovery in C-JEPA.

\begin{table}[t]
\centering
\caption{Ablation of FP supervision targets for C-JEPA W4A32 planning on Push-T.}
\vspace{-8pt}
\label{tab:supervision_ablation}
\begin{tabular}{@{}cc@{}}
\toprule
Supervision setting & Success (\%) \\
\midrule
Full joint supervision & \textbf{85.0} \\
Without direct state loss & 78.0 \\
Without future visual loss & 77.3 \\
Without future proprioceptive loss & 84.7 \\
\bottomrule
\vspace{-15pt}
\end{tabular}
\end{table}

\subsection{Additional Tasks and Resource Benefits}
\label{sec:recovery_cross_model}

\Cref{tab:task_extension} extends the evaluation beyond Push-T. Group-wise calibration restores FP-level success for DDP-WM on Wall and PointMaze and for LeWM on Two-Room. QuantWM~\cite{fu2026quantwm} reports a similar recovery for DINO-WM on Wall using OmniQuant with 128-weight groups (G128). On OGBench Cube~\cite{park2025ogbench}, LeWM retains its FP success rate of 70.0\% under uncalibrated W4 RTN. On DROID~\cite{khazatsky2024droid}, calibration raises the offline Action Score of V-JEPA2-AC~\cite{assran2025vjepa2} from 36.48 to 40.51, compared with 40.07 for FP.

\begin{table}[t]
\centering

\caption{W4 performance on additional tasks.
Simulated tasks report closed-loop success (\%);
DROID reports offline action scores.
Cube uses 50 paired episodes with uncalibrated
per-channel RTN; DINO-WM Wall follows the
50-iteration setting of QuantWM~\cite{fu2026quantwm}.}
\vspace{-8pt}
\label{tab:task_extension}
\begin{tabular}{@{}cccc@{}}
\toprule
Task / Model & FP & W4 & Recovered \\
\midrule
Wall / DDP-WM      & 94.0  & 36.0 & \textbf{94.0} \\
PointMaze / DDP-WM & 100.0 & 46.0 & \textbf{100.0} \\
Two-Room / LeWM    & 86.0  & 74.0 & \textbf{86.0} \\
Wall / DINO-WM     & 94.0  & 42.0 & \textbf{94.0} \\
Cube / LeWM        & 70.0  & \textbf{70.0} & --   \\
\midrule
DROID / V-JEPA2-AC & 40.07 & 36.48 & \textbf{40.51} \\
\bottomrule
\vspace{-10pt}
\end{tabular}
\end{table}

\begin{figure*}[t]
    \centering
    \includegraphics[width=\textwidth]{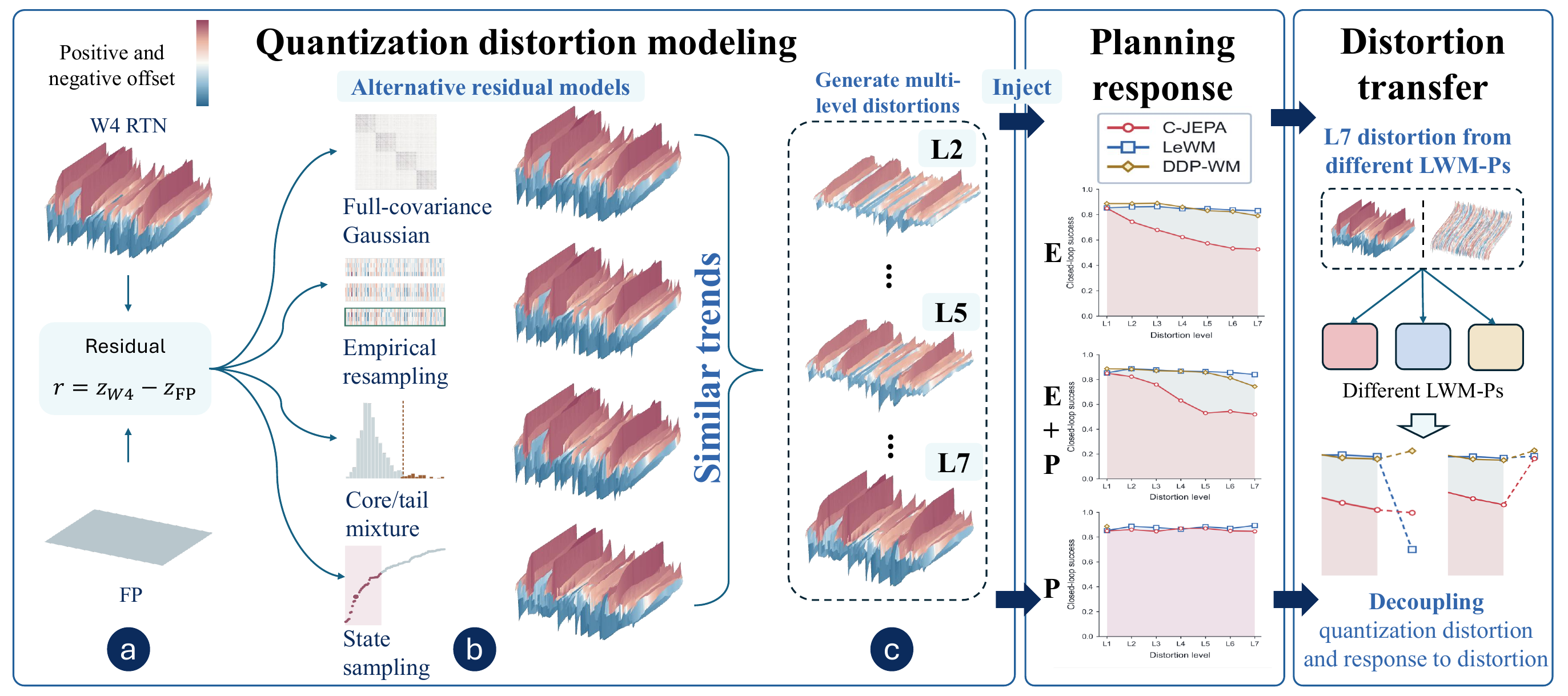}
    \vspace{-20pt}
\caption{Overview of the controlled distortion analysis.
(a) Quantization-induced latent residuals are measured
from paired FP and W4 representations at the encoder
and predictor interfaces.
(b) The measured residuals are modeled using
different distributional approximations, including
full-covariance Gaussian modeling, empirical
resampling, core--tail mixtures, and state-conditioned
sampling.
(c) The modeled residuals are progressively injected
at different strengths and interfaces while keeping
the receiving world model and planner fixed.
The resulting planning responses reveal how task
success changes with distortion strength and location.
Finally, transferring residuals across LWM-Ps
allows us to distinguish the distortion generated
by quantization from the receiving model's
sensitivity to that distortion.}
    \label{fig:distortion_workflow}
    \vspace{-10pt}
\end{figure*}

Packed W4 weights also reduce deployment costs (\Cref{tab:quantization_resource}). Across five LWM-Ps, model storage decreases by 77.0--84.0\%, GPU storage occupied by packed weight tensors by 77.1--84.0\%, and measured host-to-device weight transfer time by 56.3--71.5\%.

\begin{table}[t]
\centering

\caption{Model storage, host-to-device (H2D) weight-transfer time, and GPU storage for FP32 and packed W4 weight tensors.}
\vspace{-8pt}
\label{tab:quantization_resource}
\setlength{\tabcolsep}{3pt}
\renewcommand{\arraystretch}{1.08}
\resizebox{\columnwidth}{!}{%
\begin{tabular}{@{}ccccccc@{}}
\toprule
& \multicolumn{2}{c}{Storage (MB)}
& \multicolumn{2}{c}{H2D (ms)}
& \multicolumn{2}{c}{GPU (MB)} \\
\cmidrule(lr){2-3}
\cmidrule(lr){4-5}
\cmidrule(l){6-7}
Model & FP32 & W4 & FP32 & W4 & FP32 & W4 \\
\midrule
C-JEPA   & 105.68 & \textbf{17.13} & 37.53 & \textbf{13.79} & 105.56 & \textbf{19.68} \\
DINO-WM  & 168.80 & \textbf{27.67} & 42.21 & \textbf{12.02} & 168.72 & \textbf{27.55} \\
LeWM     & 72.29  & \textbf{16.66} & 24.61 & \textbf{10.75} & 72.17  & \textbf{16.54} \\
DDP-WM   & 186.99 & \textbf{29.99} & 49.68 & \textbf{15.16} & 186.88 & \textbf{29.84} \\
Scope-WM & 174.37 & \textbf{32.45} & 44.52 & \textbf{14.48} & 174.24 & \textbf{32.31} \\
\bottomrule
\vspace{-20pt}
\end{tabular}%
}
\end{table}

%% file: sec/5_analysis.tex
\section{Disentangling Quantization Distortion and Planning Sensitivity}
\label{sec:controlled_distortions}
Closed-loop W4 success reflects both the latent distortions produced by quantization and the planning system's response to them. We examine these factors through controlled residual injection and cross-model transfer, following the recovery experiments in \Cref{sec:recovery}.

\subsection{Controlled Residual Interventions}
\label{sec:distortion_modeling}
We measure quantization-induced differences in encoded states and predicted trajectories on matched inputs. For a quantized model $Q$, reference $B$, input $u$, and interface $\ell$, the residual is
\begin{equation}
    r_\ell(u)=z_\ell^{Q}(u)-z_\ell^{B}(u),
    \label{eq:quantization_residual}
\end{equation}
where $B$ is FP or a fixed calibrated reference. Because closed-loop planning encounters new observations, we fit resamplable Gaussian, empirical, core--tail mixture (separating typical and extreme residuals), and state-conditioned residual models (\Cref{fig:distortion_workflow}). Our state-conditioned sampler selects residuals from nearby normalized reference states, with separate banks for current and goal observations.

The measured residuals exhibit substantial state dependence. In an independent residual analysis, averaging the residuals of 32 neighboring reference states reduces estimation error by 37.3\% for current observations and 38.5\% for goals, relative to a global-mean predictor. This motivates conditioning residual sampling on the current latent state rather than modeling quantization errors solely through global distribution statistics.

For a fixed receiving model $R$, we inject sampled residuals at selected encoder or predictor interfaces:
\begin{equation}
    \tilde z_\ell^{R}(u;\beta)
    =z_\ell^{R}(u)+\beta\hat r_\ell(u),
    \label{eq:controlled_distortion}
\end{equation}
where $\beta$ controls distortion strength. Model weights and the planner remain fixed. Encoder interventions alter current and goal representations, while predictor interventions alter imagined trajectories.

\begin{figure}[t]
    \centering
    \includegraphics[width=\columnwidth]{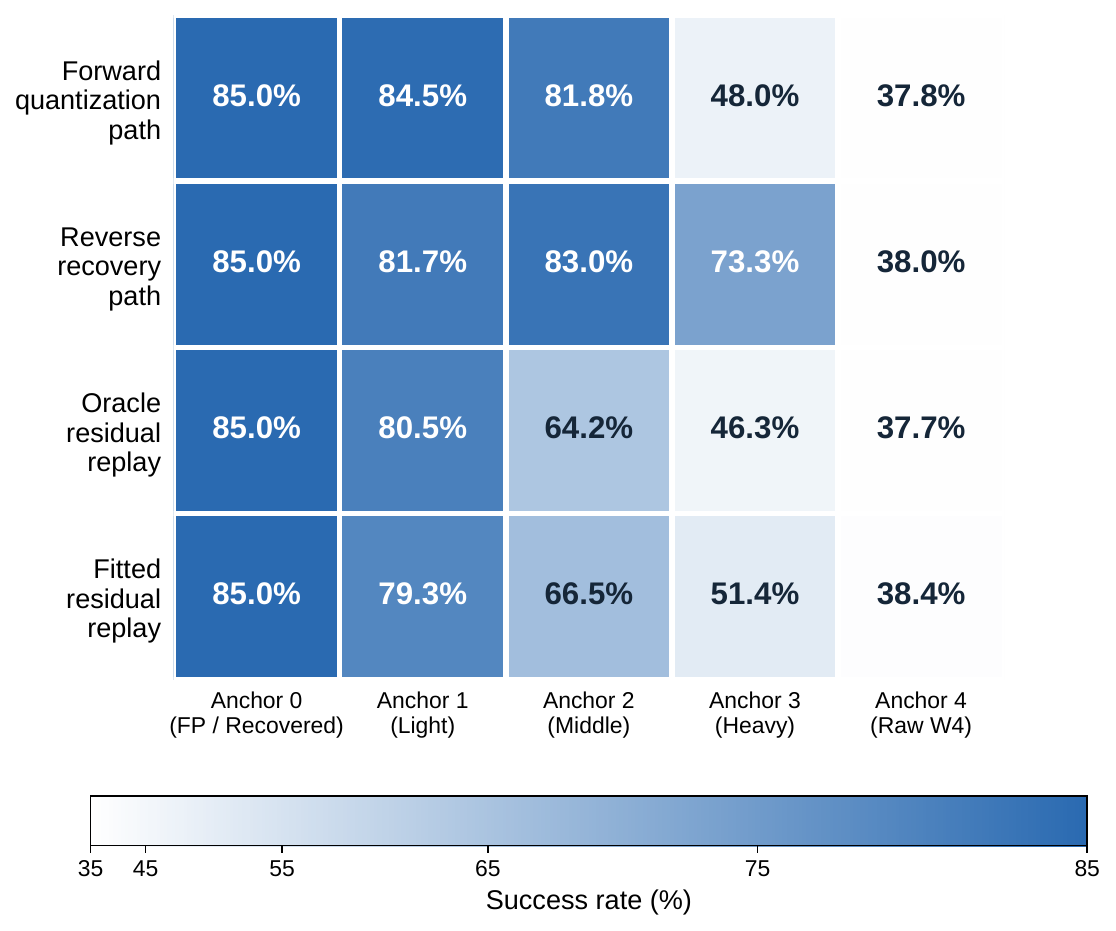}
    \vspace{-20pt}
    \caption{C-JEPA success along forward quantization, calibration, and residual-replay paths. Measured (Oracle) and fitted residual replay exhibit similar aggregate trends across five aligned strengths under closed-loop evaluation.}
    \label{fig:distortion_trends}
    \vspace{-10pt}
\end{figure}

Fitted and measured-residual replay differ by only 1.8 percentage points in mean C-JEPA success across five tested strengths (\Cref{fig:distortion_trends}). Their similar degradation trends support the use of state-conditioned residual sampling for controlled planning-response experiments.

\subsection{Distortion Generation and Response}
\label{sec:cross_model_distortion}
We contrast direct W4 quantization with the responses of fixed FP receivers to transferred encoder residuals (\Cref{tab:cross_model_distortion}). Residuals from C-JEPA and LeWM are mapped across latent spaces by semi-orthogonal projections and normalized using receiver-side FP statistics. Each receiver retains its original world model and CEM planner.

\begin{table}[t]
\centering
\caption{Push-T closed-loop success (\%) under direct W4 quantization or encoder-residual transfer into FP receivers. Each condition uses 300 episodes.}
\vspace{-8pt}
\label{tab:cross_model_distortion}
\setlength{\tabcolsep}{3pt}
\begin{tabular}{@{}ccccc@{}}
\toprule
& \multicolumn{2}{c}{Model weights}
& \multicolumn{2}{c}{Residual transfer} \\
\cmidrule(lr){2-3}\cmidrule(l){4-5}
Model & FP & W4 RTN & C-JEPA & LeWM \\
\midrule
C-JEPA & 85.0 & 37.7 & 51.7 & 83.3 \\
LeWM   & 85.3 & 82.3 & \textbf{30.7} & 85.0 \\
DDP-WM & 88.7 & 40.3 & \textbf{87.0} & -- \\
\bottomrule
\vspace{-20pt}
\end{tabular}
\end{table}

\begin{figure*}[t]
    \centering
    \includegraphics[width=\textwidth]{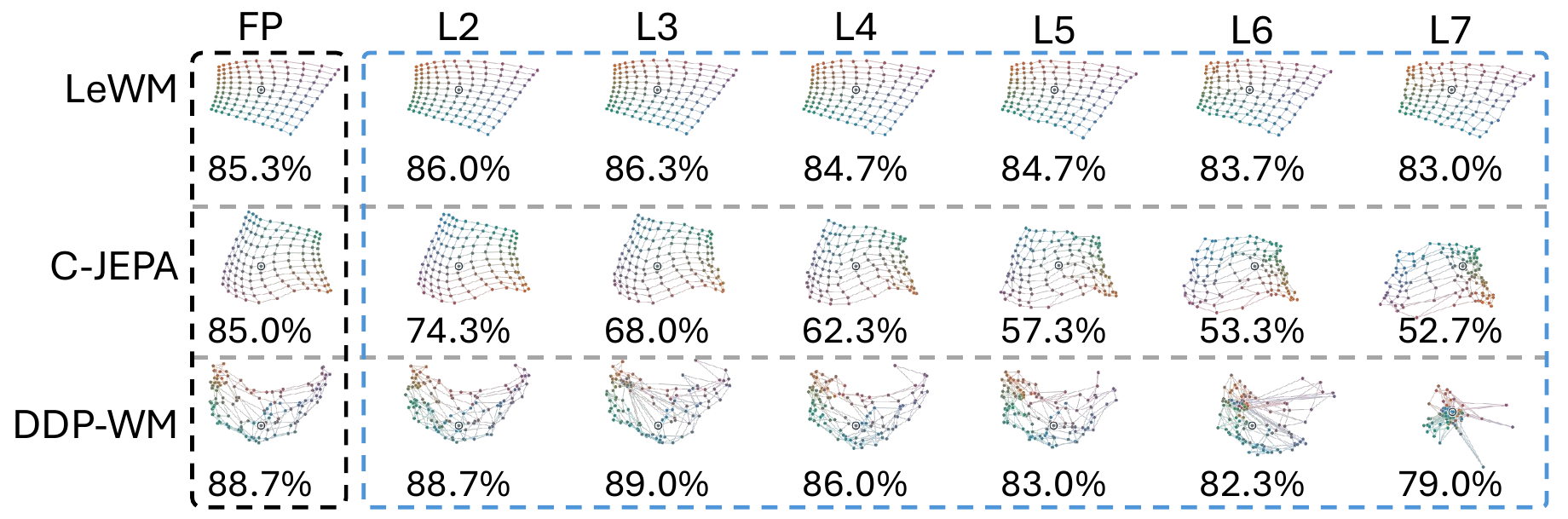}
    \vspace{-20pt}
    \caption{Local action-to-latent mappings under increasing encoder-side distortions. Each panel shows horizon-5 predictions for 121 matched actions projected onto a shared principal component analysis (PCA) basis fitted to FP predictions for each model. FP is the unperturbed baseline; L2--L7 are increasing, model-specific distortion levels. Percentages denote closed-loop success over 300 Push-T episodes per condition, enabling comparison of response geometry and task performance.}
    \label{fig:multi_model_distortion}
    \vspace{-10pt}
\end{figure*}

The comparison reveals a reversal of the apparent robustness ranking under W4. LeWM retains 82.3\% success after direct W4 quantization but falls to 30.7\% under C-JEPA-derived encoder perturbations. Conversely, DDP-WM falls from 88.7\% to 40.3\% under full W4 quantization yet retains 87.0\% under the same residual donor and 86.0\% when the transferred distortion reaches both encoder and predictor interfaces. Meanwhile, LeWM-derived residuals have much smaller root-mean-square (RMS) amplitudes relative to FP latent variation (approximately 0.082 versus 1.07 for C-JEPA), leaving the tested receivers near their FP baselines. These contrasts show why direct W4 success does not rank sensitivity to controlled latent disturbances: the distortion produced and the receiver's response both shape planning performance.

Holding the residual donor fixed further exposes differences on the receiver side. Under C-JEPA-derived encoder perturbations, closed-loop success decreases by 33.3, 54.7, and only 1.7 percentage points for C-JEPA, LeWM, and DDP-WM, respectively. Thus, a common residual source produces markedly different planning responses across receiving systems. Conversely, replacing C-JEPA residuals with the much smaller LeWM-derived residuals leaves both tested receivers close to their FP baselines. Varying the donor and receiver separately therefore exposes two distinct sources of variation that are combined in conventional W4 evaluation.

\subsection{Location-Dependent Planning Responses}
\label{sec:exposure_sensitivity}

We next examine each model's response to progressively
stronger model-specific quantization residuals, focusing
on encoder-side distortions and the effects of intervention
location on planning performance. \Cref{fig:multi_model_distortion} projects 121 matched horizon-5 predictions from a fixed Push-T input onto each model's FP latent basis. Under progressively stronger perturbations, C-JEPA's local action-response grid deforms and contracts as its success falls from 85.0\% to 52.7\%. LeWM preserves most of its local structure, with success changing only from 85.3\% to 83.0\%. DDP-WM contracts noticeably at high strengths while retaining 79.0\% from its 88.7\% FP baseline.

Intervention location further changes the planning response. At the highest tested strength, C-JEPA retains 84.7\% success with predictor-only distortion but falls to 52.0\% when encoder and predictor are perturbed together. DDP-WM likewise retains 79.0\% with encoder-only perturbation and 74.3\% with dual-interface perturbation. Scope-WM K32 extends the observed sensitivity range, falling from 76.3\% to 15.7\% under its strongest encoder perturbation. For C-JEPA, encoder-only perturbation reduces success to 52.7\%, nearly matching the 52.0\% under dual-interface injection, whereas predictor-only perturbation retains 84.7\%. Most of the tested dual-interface degradation therefore already appears at the encoder interface. This interface determines the current and goal representations used to evaluate all candidate trajectories, allowing representation distortions to alter candidate comparisons throughout planning. The different latent response patterns and location-dependent losses demonstrate that perturbation strength alone does not capture planning sensitivity.

\begin{figure}[t]
    \centering
    \includegraphics[width=\columnwidth]{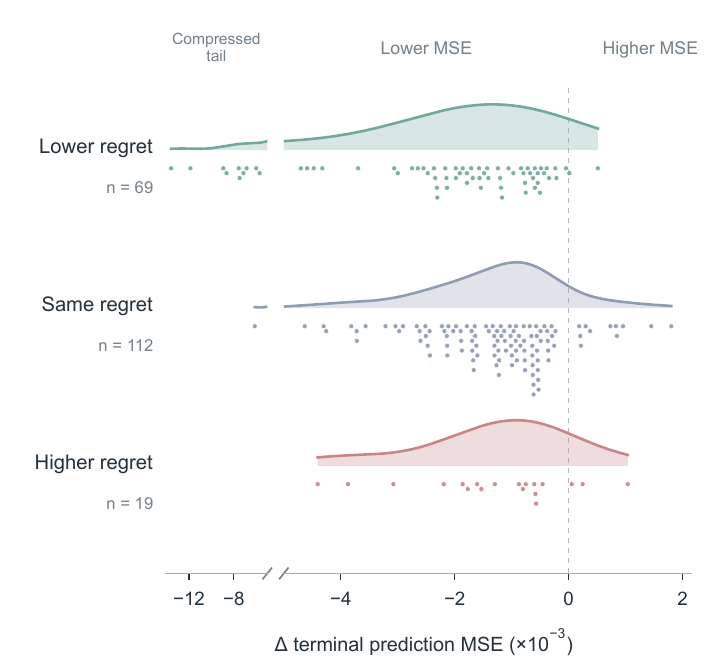}
    \vspace{-20pt}
\caption{Distribution of terminal prediction MSE changes
across 200 C-JEPA candidate pools, grouped by changes
in FP-referenced selection regret. Dots represent
individual pools, curves show group densities, and the
dashed line marks zero MSE change; negative values
indicate reductions.}
    \label{fig:prediction_decision}
\end{figure}

\subsection{Prediction Fidelity versus Decision Fidelity}
\label{sec:prediction_decision_fidelity}
Average prediction error offers another incomplete view of
planning degradation when considered in isolation.
\Cref{tab:mse_planning} compares normalized five-step
prediction error with closed-loop success under actual W4
quantization and controlled encoder perturbations. Each
success pair reports the FP baseline followed by the
success under the listed condition. For a fixed candidate
pool, FP-referenced selection regret is the FP cost of the
model-selected action minus the minimum FP cost in that pool.

\begin{table}[t]
\centering
\caption{FP-referenced five-step prediction normalized mean
squared error (NMSE) and Push-T success (\%). Each success
entry reports the FP baseline followed by the success under
the listed condition. PLDM uses 100 episodes; other models
use 300.}

\label{tab:mse_planning}
\setlength{\tabcolsep}{3pt}
\renewcommand{\arraystretch}{1.05}
\begin{tabular}{@{}cccc@{}}
\toprule
Model & Condition & NMSE & Success (\%) \\
\midrule
C-JEPA & G64 RTN       & 0.42357 & 85.0 / 37.7 \\
LeWM   & W4            & 0.01914 & 85.3 / 82.3 \\
PLDM   & W4            & 0.01364 & 71.0 / 67.0 \\
PLDM   & Extended W4   & 0.05745 & 71.0 / 67.0 \\
DDP-WM & W4            & 1.75943 & 88.7 / 40.3 \\
\midrule
LeWM   & Encoder L7    & 0.00648 & 85.3 / 83.0 \\
DDP-WM & Encoder L7    & 0.74379 & 88.7 / 79.0 \\
\bottomrule

\end{tabular}
\end{table}

In PLDM, extending W4 quantization from the encoder and predictor to the action-encoding and projection-head weights increases prediction NMSE by 4.2 times, while aggregate success remains 67.0\%. Error structure also matters within a fixed model: in C-JEPA, covariance-matched Gaussian and state-conditioned residuals produce five-step prediction errors differing by approximately 1\%, yet their success rates differ by 5.7 percentage points. Thus similar aggregate prediction errors can accompany different planning outcomes, and larger errors need not produce proportional success losses.

To examine the relationship between prediction and
action selection, we evaluate two C-JEPA calibration
checkpoints on 200 fixed candidate pools
(\Cref{fig:prediction_decision}). Global FP supervision
reduces horizon-5 prediction MSE and FP-referenced
selection regret by 33.8\% and 57.1\%, respectively,
while improving Top-1 agreement with FP's best candidate
from 51.5\% to 70.0\% and Top-30 overlap with FP's
30 lowest-cost candidates from 83.7\% to 90.6\%.
Notably, 16 of the 19 pools with increased selection
regret still exhibit reduced prediction error, showing
that improved prediction fidelity does not guarantee
better action selection.

This distinction follows from how CEM compares candidate actions. A common shift in candidate costs need not change their ordering, whereas candidate-dependent errors can reverse the preference between closely scored actions. On the fixed pools, FP supervision reduces the pairwise ranking reversal rate from 4.58\% to 2.41\%, while decreasing the error in the FP top-two cost gap from 0.00279 to 0.00197. Notably, the calibrated model reaches the FP closed-loop success rate of 85.0\% despite matching the FP Top-1 selection in only 70.0\% of diagnostic pools. Successful planning therefore need not reproduce every FP action choice, while the relative ordering and quality of candidate actions remain important.

%% file: sec/6_Conclusion.tex
\section{Conclusion}
\label{sec:conclusion}

In this work, we investigated planning recovery in low-bit quantized LWM-Ps across diverse architectures and challenging tasks.
Our experiments showed that combining local reconstruction
with global FP supervision can restore W4 planning
performance to near-FP levels.
Through controlled distortion injection and cross-model
transfer, we further identified error location,
quantization-induced distortion profiles, and model-specific
planning robustness as key factors shaping task success.
Notably, degradation under direct W4 quantization does
not necessarily reflect a model's tolerance to controlled
latent distortions.
We hope these findings will inform future studies of planning recovery and robustness in quantized world models.

%% file: sec/X_suppl.tex
\clearpage
\appendix
\setcounter{section}{0}
\setcounter{table}{0}
\setcounter{figure}{0}
\setcounter{equation}{0}
\renewcommand{\thetable}{S\arabic{table}}
\renewcommand{\thefigure}{S\arabic{figure}}
\renewcommand{\theequation}{S\arabic{equation}}
\maketitlesupplementary
\captionsetup[table]{justification=justified,singlelinecheck=false}
\captionsetup[figure]{justification=justified,singlelinecheck=false}

\section{Evaluation Protocols}
\label{app:protocol}

\subsection{Tasks and calibration data}
The principal Push-T comparisons use 300 tasks per condition. C-JEPA, LeWM, and DDP-WM share the expert-trajectory tasks used in the controlled distortion analysis, with FP success counts of 255, 256, and 266. These tasks are sampled from expert trajectories, as in LeWM. DINO-WM and Scope-WM recovery use 300 draws from the Push-T noise-validation data, comprising 268 distinct windows from 21 trajectories.

Calibration uses 100 fitting and 20 development trajectories from the models' training data. Matched comparisons use the same windows, action pools, and frozen normalization statistics, with supervision objectives specified in Section~\ref{app:calibration}. Development data are used to monitor calibration losses.

\subsection{Native planning}
Within each model, FP and quantized conditions share task order, initial states, goals, preprocessing, planner costs, and success criteria. Table~\ref{tab:splanner} gives the principal CEM budgets. Candidate scoring uses the complete population in each call.

\begin{table}[!htbp]
\centering\small
\caption{Native Push-T planning budgets. $H$ is the latent prediction horizon.}
\label{tab:splanner}
\setlength{\tabcolsep}{4pt}
\begin{tabular}{@{}ccccc@{}}
\toprule
Model & $H$ & Candidates & Elites & Iterations \\
\midrule
C-JEPA & 5 & 300 & 30 & 30 \\
LeWM & 5 & 300 & 30 & 30 \\
DDP-WM & 5 & 300 & 30 & 30 \\
DINO-WM & 5 & 300 & 30 & 10 \\
Scope-WM K32 & 5 & 300/100 & 30/10 & 10 \\
\bottomrule
\end{tabular}
\end{table}

C-JEPA and LeWM execute up to two 25-control blocks; DDP-WM uses three blocks with its native sparse cost and stopping rule. DINO-WM allows 15 MPC rounds and removes successful tasks from subsequent scoring. Scope-WM uses 300 candidates initially and 100 thereafter, retaining its top-K32 selector and foreground/background updates. C-JEPA's goal is the frame at offset 24 under its native 25-control half-open interval.

The additional simulated tasks use 50 paired tasks per condition and retain the native observations, normalization, goal construction, and termination. The DINO-WM Wall row of Table~\ref{tab:task_extension} reports the FP, W4, and OmniQuant G128 results of QuantWM~\cite{fu2026quantwm} under its 50-iteration protocol. PLDM Push-T uses 100 tasks from 96 source episodes; extended W4 additionally covers action-encoding and projection-head weights.

\subsection{Uncertainty and hardware}
Paired bootstrap intervals resample source trajectories, retaining all windows and planner seeds within each sampled trajectory. We use 20,000 resamples for closed-loop comparisons, 10,000 state-level resamples for fixed-pool diagnostics, and 2,000 trajectory-level resamples for DROID. GPU experiments run on NVIDIA GeForce RTX 2080 Ti cards with memory upgraded to 22 GB, in the native model environments, including PyTorch 2.3.0/CUDA 12.1. The common-runtime fixed-pool analysis uses PyTorch 2.13.0 on CPU.

\section{Calibration Methods}
\label{app:calibration}

\subsection{Quantized weights and trainable parameters}
The main setting is W4A32. G64 partitions each output row of an eligible weight matrix into consecutive groups of 64 elements; per-channel quantization uses the whole row as one group. Convolutional kernels are flattened along their input and spatial axes. Symmetric RTN initializes each group with scale $s_g=\max_{j\in g}|w_j|/7$ and integer codes in $[-7,7]$.

Group calibration learns a log-scale $\alpha_g$ and a rounding variable $v_j$ for each weight:
\begin{equation}
\widetilde w_j=s_g^{\mathrm{ref}}e^{\alpha_g}
\operatorname{clip}_{[-7,7]}\!\left(b_j+\mathbf1[v_j>0]\right).
\label{eq:squantizer}
\end{equation}
Here $b_j$ is the lower code of the initializer's adjacent-code pair. The forward pass uses hard codes, while a sigmoid straight-through estimator supplies rounding gradients. Continuing from a calibrated checkpoint preserves its scales, hard-code choices, and admissible rounding grid at initialization. Original FP weights, biases, and modules outside the quantization target set remain fixed.

C-JEPA quantizes encoder projections, packed attention weights, Slot Attention GRU/update weights, and predictor projections. Thus its calibration can adjust features before and within object aggregation. The other architectures target their native visual encoders and predictors; auxiliary modules retain their native precision.

\subsection{Observed states and recursive FP targets}
Each calibration window supplies six snapshots at offsets $0,5,\ldots,25$ and eight five-step action sequences: the recorded sequence and seven fixed Gaussian alternatives in normalized action coordinates. Each latent action concatenates five primitive two-dimensional controls. These eight training sequences are separate from CEM's planning population.

The frozen FP teacher encodes each of the six snapshots independently. It also starts from its own initial encoding and recursively predicts under every action sequence. The student independently encodes the same snapshots and rolls out from its own initial state. Future images supervise the direct state term; recursive student predictions receive the initial observation and actions. FP targets and training-side feature normalization stay fixed throughout fitting.

\subsection{Local and global supervision}
\label{app:global_loss}
Local reconstruction runs each quantized block on inputs captured from the FP model and matches that block's FP output. Repeated calls are averaged within each block, and blocks receive equal weight:
\begin{equation}
\mathcal L_{\mathrm{local}}
=\frac1{|\mathcal B|}\sum_{b\in\mathcal B}
\operatorname{mean}_{j}
\operatorname{MSE}\!\left(f_b^Q(x_{b,j}^{\mathrm{FP}}),
y_{b,j}^{\mathrm{FP}}\right).
\end{equation}
C-JEPA uses 24 native reconstruction units spanning the encoder, aggregation, and predictor. Each output loss updates the quantizers inside its block.

Global FP supervision instead evaluates the assembled student. Let $z_j$ denote an encoded visual state and $(\hat z_h^i,\hat p_h^i)$ a recursive visual/proprioceptive prediction for action sequence $i$ at horizon $h$. The objective is
\begin{equation}
\begin{aligned}
\mathcal L_{\mathrm{global}}={}&
\operatorname{mean}_{j}\mathcal D(z_j^Q,z_j^{\mathrm{FP}})\\
&+\operatorname{mean}_{i,h}\mathcal D(\hat z_h^{Q,i},
\hat z_h^{\mathrm{FP},i})\\
&+\operatorname{mean}_{i,h}\mathcal D_p(\hat p_h^{Q,i},
\hat p_h^{\mathrm{FP},i}).
\end{aligned}
\label{eq:sglobal}
\end{equation}
The three terms have equal coefficients. $\mathcal D$ is MSE after channelwise normalization by FP training standard deviations. For C-JEPA's four unordered 128-dimensional slots, it minimizes this distance over slot permutations independently for each state or candidate/horizon pair. Ordered patch and vector representations retain their native correspondence. $\mathcal D_p$ is normalized proprioception MSE and is omitted for architectures without that output. DDP-WM additionally aligns its selector logits with the FP selector.

The state term directly updates encoder and aggregation quantizers. The prediction terms differentiate through all student rollout steps and its initial encoding, coupling encoder and predictor calibration. FP predictions remain detached. All action branches, including the recorded one, use FP recursive predictions as future targets.

For the matched C-JEPA local/global comparison, both branches start from the same hard G64 checkpoint. This common initialization was fitted for 500 updates using one initial observation, 64 fixed Gaussian action sequences per window, and FP recursive prediction targets. Each comparison branch then receives 550 updates, with identical data order and AdamW settings, using either local reconstruction or Equation~\eqref{eq:sglobal}. Each update consumes one fitting window and its complete training action pool. The exported checkpoint uses the final hard codes and scales. Supervision ablations remove the corresponding loss term; scale-only fitting holds all rounding decisions fixed.

\subsection{PTQ initializations}
The PTQ comparison uses the same C-JEPA weight coverage and G64 layout. Each initialization is followed by 550 updates of global FP supervision, with its method-specific variables:

\paragraph{RTN and OMSE.}
RTN uses group absmax scales. OMSE searches smaller clipping ranges and selects the scale minimizing groupwise weight MSE. Global calibration then learns group scales and rounding choices.

\paragraph{AWQ and SmoothQuant.}
Both use an equivalent input-channel rescaling,
\begin{equation}
\widetilde W=Q_{64}(WD)D^{-1},
\end{equation}
with diagonal $D$. AWQ selects activation-aware scales and clipping ranges by affine-output reconstruction on fitting inputs. SmoothQuant balances input and weight maxima with smoothing exponent $0.5$. In our W4A32 setting, activations remain FP32, so SmoothQuant acts only as an equivalent weight transformation. The subsequent global stage fixes $D$ and updates the quantizer's scales and rounding variables.

\paragraph{OmniQuant.}
Learnable equivalent transformations and weight clipping are fitted sequentially over the 24 native blocks. A block receives inputs from the calibrated prefix and matches the complete FP teacher's block output. Equivalent transformations include reciprocal input/weight scaling and Q/K scaling; translation stays zero. Local fitting uses 20 epochs per block. Global calibration subsequently optimizes these transformation and clipping variables through the assembled model.

\paragraph{G64 block MSE.}
The local and global conditions use the common G64 initialization described above. Their difference is the supervision location: native block outputs versus encoded states and recursive predictions. The grouped scale/rounding parameterization is shared.

\subsection{Interface compensation}
\label{app:interface_correction}
Interface compensation freezes the quantized backbone and learns residual corrections at its encoder and predictor outputs. The encoder branch projects normalized tokens to a low-rank hidden space, combines each token with pooled state context, applies SiLU, and projects back to the native latent dimension. The predictor branch conditions its correction on the current state, the proposed latent increment, and native action/proprioception inputs. These inputs modulate the low-rank hidden features before the output projection. An auxiliary head corrects proprioception where present. Output projections initialize to zero, so fitting begins with the original quantized model.

The encoder correction is applied to observations and goals. Predictor corrections enter every recursive step and feed the next prediction. In C-JEPA, encoder compensation acts on the four visual slots \emph{after} Slot Attention. C-JEPA, DINO-WM, and DDP-WM use rank 32; LeWM and Scope-WM use rank 16.

State supervision aligns the same six independently encoded snapshots used in global quantizer calibration. Future supervision uses FP encodings of recorded observations for the executed action branch and FP recursive predictions for unexecuted alternatives. On the latter, we match both the candidate-mean feature and candidate-dependent deviations from that mean:
\begin{equation}
\mathcal L_{\mathrm{contrast}}=
\operatorname{mean}_h
\frac{\operatorname{mean}_{i,f}(d_{ihf}^Q-d_{ihf}^T)^2}
{\max(\operatorname{mean}_{i,f}(d_{ihf}^T)^2,\tau)},
\label{eq:scontrast}
\end{equation}
where $d_{ih}=F_{ih}-\bar F_h$, $\bar F_h$ averages eligible candidates, $f$ indexes feature coordinates, and $\tau$ is a fitting-data variance floor. Only horizons with at least two eligible candidates contribute. Ordered representations use normalized native features; C-JEPA uses a fixed permutation-invariant summary of its normalized visual slots.

The visual pool loss combines observed-future alignment with FP candidate-mean and contrast terms, weighted $1$, $0.05$, and $0.25$. The full objective adds six-frame state alignment, observed/future proprioception alignment, and output-projection regularization. Gradients traverse the frozen native dynamics to train both compensation branches. The main compensation recipe uses 50 state-only warmup updates followed by 500 joint updates and development-based checkpoint selection. The controlled C-JEPA location comparison uses 550 joint updates from the same zero-output initialization.

\section{Residual Construction and Injection}
\label{app:residual}

\subsection{Paired residual banks}
For each interface, a training bank stores reference states and their paired residuals,
\begin{equation}
\mathcal B=\{(c_j,r_j)\}_{j=1}^n,\qquad
c_j=\operatorname{vec}(z_j^B),\quad
r_j=\operatorname{vec}(z_j^Q-z_j^B).
\end{equation}
Both models receive identical images, actions, and initialization seeds. C-JEPA/LeWM encoder-strength curves use encoder RTN minus a jointly calibrated reference, while retaining that reference's predictor. Full-model replay, cross-model transfer, and the comparison at matched prediction error use G64 RTN minus FP. Because LeWM's direct-W4 results use native per-output-channel RTN, Section~\ref{app:transfer_controls} additionally transfers LeWM residuals generated with that quantizer.

Encoder banks contain six snapshots from each of 100 fitting windows. C-JEPA keeps separate current/goal banks of $4\times128$ residual tensors; LeWM uses 192-dimensional vectors. DDP-WM and Scope-WM retain complete $196\times384$ visual tensors. Native slot or patch order is preserved. Predictor residuals compare FP and quantized predictions given the same FP history and actions. Banks are separated by native call and output horizon. C-JEPA/LeWM use all eight training actions; the larger DDP-WM/Scope-WM banks retain the recorded action and one fixed alternative per window.

\subsection{Fitted residual distributions}
Let $\mu$ and $\Sigma$ be the empirical mean and population covariance of a bank. We compare five samplers:

\paragraph{Isotropic Gaussian.}
Draw zero-mean Gaussian noise with variance $\mathbb E\|r\|_2^2/d$, matching the bank's average squared residual magnitude.

\paragraph{Covariance-matched Gaussian.}
Draw $\hat r\sim\mathcal N(\mu,\Sigma)$, retaining correlations across channels and tokens. Sampling from the centered residual matrix avoids explicitly storing a large dense covariance.

\paragraph{Empirical resampling.}
Draw one complete residual tensor uniformly from the bank, retaining its joint coordinate structure.

\paragraph{Core--tail mixture.}
Split residuals at the training 95th percentile of tensor norm. Fit a full-covariance Gaussian to each subset and sample the components according to their observed frequencies.

\paragraph{State-conditioned residuals.}
Standardize reference-state coordinates using fitting-data means and standard deviations. Find the 32 nearest bank states to the receiver's current interface output and uniformly draw one of their associated residual tensors. Distances for large predictor tensors use a fixed subset of 1,536 coordinates; each sampled residual retains all coordinates. Residuals are injected in native latent units. Sampling uses a separate, deterministic state-keyed random-number stream, so repeated interface values receive the same residual independently of CEM sampling.

\subsection{Injection interfaces and strength}
Encoder injection adds $\beta_E\hat r_E$ to current and goal representations before candidate rollouts. C-JEPA intervenes after Slot Attention; LeWM intervenes on its encoded state vector; DDP-WM and Scope-WM intervene on the visual output while preserving proprioception and their native selectors. At each MPC cycle, the new observation is encoded and perturbed again.

Predictor injection acts after each native prediction and before its use in the next-step history:
\begin{equation}
q_{h+1}=P_R(\widetilde{\mathcal H}_h,a_h),\qquad
\widetilde z_{h+1}=q_{h+1}+\beta_P\hat r_{P,h}(q_{h+1}).
\end{equation}
Visual-interface experiments perturb visual coordinates, retaining native action/proprioception handling. Full-model replay also includes C-JEPA's nonvisual predictor outputs. Identical current states expanded over candidates receive the same encoder perturbation; distinct predicted states query their own predictor residuals. The receiving model, cost function, CEM, and MPC execution remain fixed.

For C-JEPA, LeWM, and DDP-WM, encoder levels L1--L7 correspond to seven training-side prediction-error targets. Let $\mathcal E(\beta)$ be five-step visual MSE after encoder injection, averaged over the fixed eight-action pools, and let $\mathcal E_{\mathrm{RTN}}$ be the error of the encoder-RTN control. We fit a nonnegative multiplier to minimize
\begin{equation}
\left|\mathcal E(\beta)-\left[\mathcal E(0)
+t\big(\mathcal E_{\mathrm{RTN}}-\mathcal E(0)\big)\right]\right|,
\end{equation}
for $t=0,1/6,\ldots,1$. Levels are thus defined relative to each model's own encoder-RTN error, so equal level indices do not imply equal perturbation amplitudes across models; LeWM's levels, in particular, are bounded by its small encoder-RTN error (Section~\ref{app:transfer_controls} compares receivers at a common relative amplitude). Scope-WM uses evenly spaced multipliers directly. Predictor-only and dual-interface curves also use $\beta=0,1/6,\ldots,1$. The action-response plots score 121 matched actions and project all conditions onto the same model-specific PCA basis fitted to FP predictions.

Measured (Oracle) replay computes $z^{\mathrm{FP}}+\lambda(z^{\mathrm{W4}}-z^{\mathrm{FP}})$ online, evaluating both branches on the receiver's same current input/history. The blended output enters subsequent recursive steps. Fitted replay replaces the measured difference with a state-conditioned bank sample. Both use the same encoder/full-predictor interfaces and five strengths in the main-paper replay comparison.

\subsection{Cross-model residual transfer}
\label{app:transfer_implementation}
Cross-model transfer draws complete empirical G64 RTN-minus-FP donor residuals and maps them into a fixed receiver. Let $\sigma_D$ and $\sigma_R$ be FP latent RMS values after subtracting the respective training mean latent. RMS averages over samples and feature coordinates. For a donor residual row vector $r$ and dimension map $A$, define
\begin{equation}
\delta_{D\to R}(r)=(rA)
\frac{\operatorname{rms}(r)}{\operatorname{rms}(rA)}
\frac{\sigma_R}{\sigma_D}.
\label{eq:stransfer}
\end{equation}
This preserves donor-relative amplitude in receiver units:
$\operatorname{rms}(\delta_{D\to R})/\sigma_R
=\operatorname{rms}(r)/\sigma_D$.

For C-JEPA and LeWM, a fixed Gaussian QR basis $Q\in\mathbb R^{512\times192}$ maps flattened C-JEPA slots to LeWM vectors; the reverse direction uses $Q^\top$. Self-transfer uses the identity. The RMS restoration in Equation~\eqref{eq:stransfer} compensates for projection-induced magnitude changes.

For C-JEPA to DDP-WM, we preserve the token/channel factorization. Fixed orthonormal-column matrices $U\in\mathbb R^{196\times4}$ and $V\in\mathbb R^{384\times128}$ map a residual $r\in\mathbb R^{4\times128}$ as
\begin{equation}
\delta_{D\to R}(r)=UrV^\top
\sqrt{\frac{196\cdot384}{4\cdot128}}
\frac{\sigma_R}{\sigma_D}.
\end{equation}
The dimension factor restores per-coordinate RMS. Predictor transfer uses the same map with separate FP scales and residual banks for each horizon. Perturbations enter the latest visual prediction before recursion.

Mapped encoder banks retain separate current/goal roles. A deterministic hash of the receiver state selects a bank entry, so repeated states receive the same transferred residual and candidate expansion reuses that perturbation. Cross-model selection uses this state hash in place of within-model nearest-neighbor sampling. Projection bases and normalization statistics remain fixed throughout evaluation. Transferred residuals retain donor-relative RMS values of approximately 1.07 for C-JEPA and 0.082 for LeWM. Section~\ref{app:transfer_controls} repeats the transfer with additional random projections, at matched relative amplitude, and with LeWM residuals from its native quantizer.

\section{Supplementary Comparisons}
\label{app:recovery}

\subsection{Quantizer freedom and compensation location}
In the main C-JEPA checkpoint, global FP supervision jointly optimizes group scales and rounding logits. It changes 402,827 group scales and 25,768,386 rounding logits, yet none of these logit changes alters a hard integer code, so all codes remain at their G64 initialization. Consistent with this, explicitly freezing rounding and fitting only scales yields 253/300, compared with 255/300 for full joint calibration. Recovery under global supervision is therefore driven mainly by scale adaptation.

Table~\ref{tab:sbridges} separates encoder and predictor compensation and tests compensation after joint calibration. The location comparison uses the common G64 initialization of Section~\ref{app:global_loss} (200/300) and a matched compensation parameter budget. This initialization is the parent of the local-reconstruction checkpoint in Table~\ref{tab:ptq_supervision}, which reaches 214/300 after 550 further updates. Encoder-only compensation recovers 22 additional tasks, compared with 6 for predictor-only compensation, and applying both yields 224/300, consistent with the encoder-dominated degradation in Section~\ref{sec:exposure_sensitivity}. Joint calibration already matches FP success (255/300), and adding compensation on top of it yields 247/300. Compensation thus helps on the weaker base but not on top of joint calibration.

\begin{table}[!tbp]
\centering\small
\caption{C-JEPA success counts out of 300 with interface compensation. Single-interface branches match the total parameter budget of both branches.}
\label{tab:sbridges}
\setlength{\tabcolsep}{4pt}
\begin{tabular}{@{}>{\centering\arraybackslash}m{2.5cm}cccc@{}}
\toprule
Quantized base & None & Encoder & Predictor & Both \\
\midrule
G64 initialization & 200 & 222 & 206 & 224 \\
Joint calibration & 255 & -- & -- & 247 \\
\bottomrule
\end{tabular}
\end{table}

\subsection{Weight precision}
\label{app:precision}
Table~\ref{tab:sbits} varies weight precision using direct G64 RTN and FP32 activations; prediction and candidate-selection metrics use shared action pools. C-JEPA retains near-FP success at W6--W8, while W5 and W4 produce substantially larger losses. Top-30 overlap falls monotonically as bit width decreases, whereas prediction MSE is not monotonic below W4: W2 has a lower MSE than W3 (0.118 vs.\ 0.446) although both fail (16 and 17/300). This provides another instance in which aggregate prediction error does not track planning outcome.

\begin{table}[!tbp]
\centering\small
\caption{C-JEPA precision sweep on 300 expert tasks. MSE is the horizon-5 prediction error and differs from the NMSE in Table~\ref{tab:mse_planning}.}
\label{tab:sbits}
\setlength{\tabcolsep}{5pt}
\begin{tabular}{@{}cccc@{}}
\toprule
Weights & Success & MSE & \shortstack{Top-30\\overlap (\%)} \\
\midrule
FP32 & 255 & 0 & 100.0 \\
W8 & 253 & 0.000345 & 93.08 \\
W7 & 257 & 0.001598 & 87.41 \\
W6 & 245 & 0.005166 & 76.44 \\
W5 & 144 & 0.017514 & 51.58 \\
W4 & 113 & 0.022303 & 42.05 \\
W3 & 17 & 0.446170 & 16.37 \\
W2 & 16 & 0.117930 & 9.64 \\
\bottomrule
\end{tabular}
\end{table}

\subsection{Residual structure at matched prediction error}
Table~\ref{tab:smatched} tests whether residual structure affects planning when prediction error is held fixed. In jointly calibrated C-JEPA, we compare covariance-matched Gaussian and state-conditioned encoder residuals, both sampled from G64 RTN-minus-FP banks. For a target fraction of the prediction-error increase from the unperturbed model to encoder RTN, both samplers' amplitudes are fitted on training windows to produce the same five-step prediction error. The match is then verified on the first or final CEM pool of the first MPC cycle and accepted when the absolute MSE difference is within 2\% of the two samplers' mean. The fractions $1/3$ on the first pool and $1/6$ on the final pool pass this check (0.52\% and 1.04\%). The matched pair at $2/3$ on the first pool fails it (8.03\%) and is therefore not evaluated in closed loop. Accepted conditions run closed-loop on 300 matched tasks.

\begin{table}[!tbp]
\centering\small
\caption{Covariance-matched Gaussian and state-conditioned encoder residuals at matched prediction error. Counts are successes out of 300. MSE gap is the absolute difference between the two samplers' prediction errors relative to their mean; only target fractions with a gap of at most 2\% are shown.}
\label{tab:smatched}
\setlength{\tabcolsep}{4pt}
\begin{tabular}{@{}ccccc@{}}
\toprule
CEM pool & \shortstack{Target\\fraction} & Gaussian & Conditioned & MSE gap \\
\midrule
First & $1/3$ & 182 & 192 & 0.52\% \\
Final & $1/6$ & 203 & 220 & 1.04\% \\
\bottomrule
\end{tabular}
\end{table}

At both matching points, state-conditioned residuals yield higher success than covariance-matched Gaussian residuals despite nearly identical prediction error: by 5.67 points on the final pool (paired 95\% interval $[0.67,10.67]$) and by 3.33 points on the first pool ($[-1.67,8.33]$).

\begin{figure*}[!t]
\centering
\includegraphics[width=\textwidth]{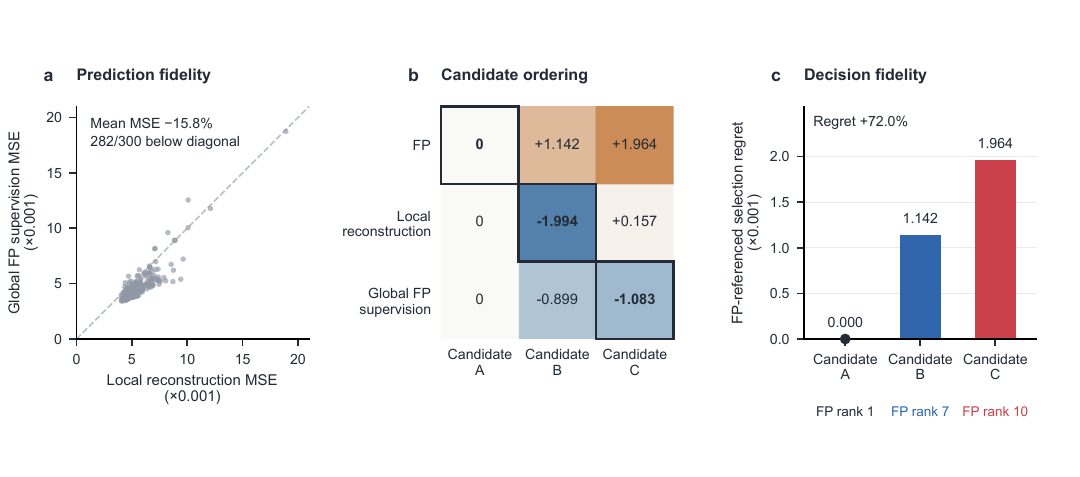}
\caption{A fixed-pool C-JEPA example linking prediction fidelity and decision fidelity. (a) Each point is one of 300 shared action sequences; points below the dashed diagonal have lower horizon-5 prediction MSE with global FP supervision than with local reconstruction. MSE sums matched visual and proprioceptive errors. (b) Native costs ($\times10^{-3}$) for candidates A, B, and C, selected by FP, local reconstruction, and global FP supervision, respectively. Each row subtracts that model's cost of candidate A, preserving its preferences; boxes mark the selected minimum. (c) FP-referenced selection regret of candidates A--C. Global FP supervision reduces mean prediction MSE but selects a candidate with higher FP cost.}
\label{fig:sfixed_pool_case}
\end{figure*}

\section{Diagnostics and Resource Measurements}
\label{app:decision}

\subsection{Prediction and candidate selection}
For the five-step NMSE comparison, errors are measured against FP predictions on the same inputs and normalized by FP latent variance. C-JEPA/DDP-WM use shared candidate pools and training-side FP variance; LeWM/PLDM use recorded segments and their corresponding FP variance. C-JEPA visual errors minimize over slot permutations; patch/vector representations retain coordinate order. Because C-JEPA and DDP-WM share the same protocol, their entries in Table~\ref{tab:mse_planning} are directly comparable: DDP-WM under encoder L7 has a higher NMSE than C-JEPA under G64 RTN (0.744 vs.\ 0.424), yet retains 79.0\% rather than 37.7\% success.

The fixed-pool study uses 50 expert state/goal pairs and four FP CEM pools per state, with 300 action sequences each. FP and both calibrated checkpoints score these same 60,000 sequences from their own encoded states in one CPU runtime. Prediction MSE includes horizon-5 visual and proprioceptive errors; native cost combines visual set matching and proprioception.

For costs $c_i^{\mathrm{FP}}$ and $c_i^Q$ in a shared pool, FP-referenced selection regret is
\begin{equation}
\operatorname{Regret}(Q)=c_{i_Q}^{\mathrm{FP}}-\min_i c_i^{\mathrm{FP}},
\qquad i_Q=\arg\min_i c_i^Q.
\end{equation}
Top-1 agreement counts identical selected candidates. Top-$k$ overlap is the intersection of the two best-$k$ sets divided by $k$. Pairwise ranking reversal counts oppositely ordered candidate pairs among FP non-ties. FP top-two gap error measures the error in the cost difference between the two best FP candidates. Pool comparisons and bootstrap resampling retain the four pools belonging to each state.

Of the 200 fixed pools, 19 show higher selection regret under global FP supervision, and 16 of these also have lower prediction MSE; across all 200 pools, global FP supervision reduces selection regret by 57.1\%. Figure~\ref{fig:sfixed_pool_case} shows the pool with the median regret increase among these 16. In this pool, mean horizon-5 prediction MSE decreases by 15.8\%, while the selected candidate ranks tenth by FP, compared with seventh for local reconstruction. The three models score the same observation, goal, and 300 action sequences.

\subsection{Open-loop image predictions}
\begin{figure*}[!t]
\centering
\includegraphics[width=\textwidth]{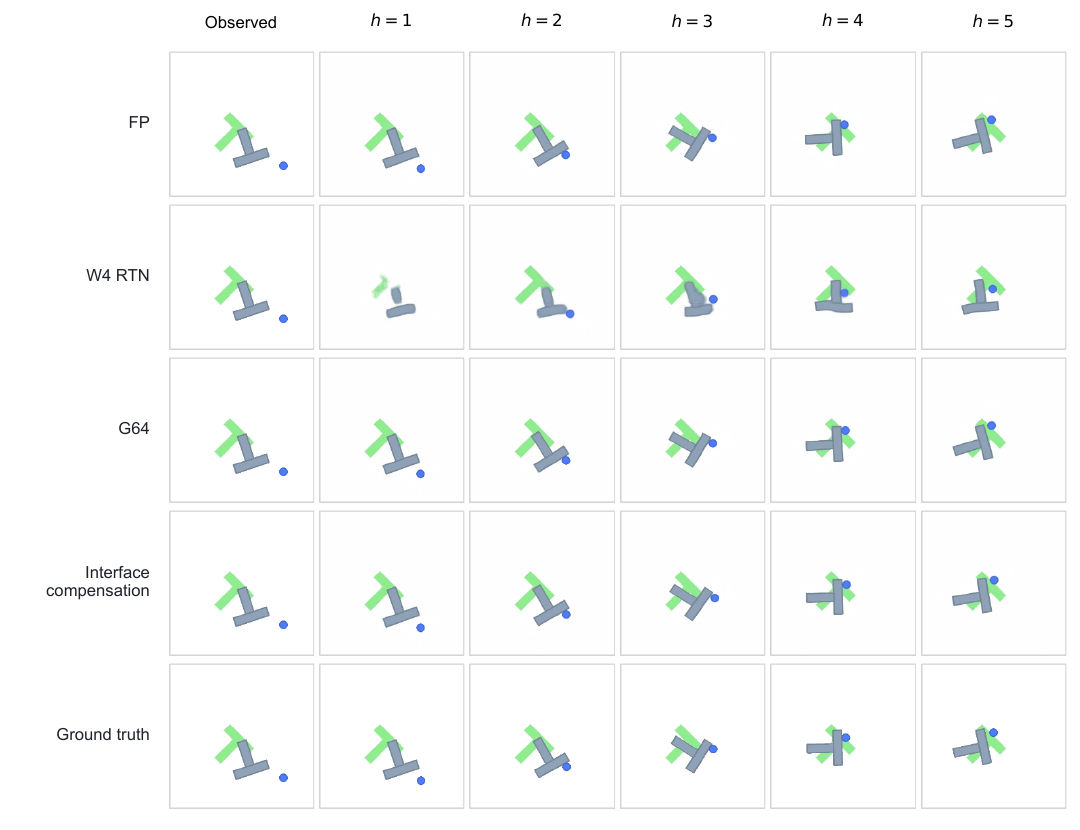}
\caption{Open-loop image predictions in DINO-WM. All models receive the same three-frame history and recorded actions. The first column repeats the last observed frame; the next five columns show recursive predictions, with five environment steps per latent step. W4 RTN uses per-output-channel symmetric quantization. G64 denotes the fitted W4 backbone; interface compensation adds the state and prediction corrections to that backbone. All predicted images use the same frozen FP VQ-VAE decoder. The bottom row shows the recorded future.}
\label{fig:sopenloop}
\end{figure*}

Figure~\ref{fig:sopenloop} visualizes DINO-WM predictions under the same recorded action sequence. W4 RTN distorts the decoded object's geometry and rotation; G64 calibration and interface compensation produce sequences closer to FP. The original frozen FP decoder renders each model's latent predictions.

In the W4 RTN row, the T shape becomes more coherent from $h=1$ to $h=5$. Table~\ref{tab:sopenloop_horizon} reports the corresponding errors for this sequence. The predictor uses a rolling three-frame context; from $h=4$ onward, this context contains only previously predicted states. This pattern is consistent with the encoder-side sensitivity observed in Section~\ref{sec:exposure_sensitivity}: once the context no longer contains encoder outputs, encoder-induced feature distortions stop entering directly, and the decoded shape becomes more coherent while errors in object position and orientation persist.

\begin{table}[!tbp]
\centering\small
\caption{Errors for the W4 RTN sequence in Figure~\ref{fig:sopenloop}. Latent MSE is relative to FP predictions; image MSE is relative to the recorded future, with RGB values in $[0,1]$.}
\label{tab:sopenloop_horizon}
\setlength{\tabcolsep}{8pt}
\begin{tabular}{@{}ccc@{}}
\toprule
Error & $h=1$ & $h=5$ \\
\midrule
Latent MSE & 1.449 & 0.473 \\
Image MSE & 0.00362 & 0.00600 \\
\bottomrule
\end{tabular}
\end{table}

\subsection{DROID Action Score}
DROID evaluation uses 64 recorded windows from 14 trajectories for V-JEPA2-AC. Each window receives one CEM planning call with horizon 3, 15 iterations, 300 candidates, 10 elites, visual L2 cost, and recursive context length 2. For window $w$, let
\begin{equation}
e_w=\sum_{j\in\{x,y,z\}}\left|
\sum_t a^{\mathrm{plan}}_{w,t,j}-
\sum_t a^{\mathrm{recorded}}_{w,t,j}\right|.
\end{equation}
The offline Action Score is
$\max(0,800(0.1-\operatorname{mean}_w e_w))$, with maximum 80. Averaging precedes truncation. It measures agreement with the recorded displacement. G64 fitting uses layerwise FP reconstruction, and the recovered score in Table~\ref{tab:task_extension} uses this fitting.

\subsection{Packed weights and memory}
\label{app:deployment}
Each byte stores two four-bit codes. Exported files also contain FP32 scales, biases, unquantized weights, and transformation buffers. All sizes are reported in decimal MB ($10^6$~bytes). Persistent GPU storage sums resident codes, scales, retained FP weights, and buffers. H2D time is the median of 30 interleaved, CUDA-synchronized transfers from CPU-ready tensors on RTX 2080 Ti.

DINO-WM packing disables its unused decoder. Shared DDP-WM weights are deduplicated; its transfer measurement uses tensors matching the exported shapes, dtypes, and byte counts. LeWM transfers its source and packed tensors. Scope-WM retains FP32 selection/background-update modules.

In closed-loop planning, W4 weights are expanded to FP32 only when used, so packing reduces persistent memory without altering planning: in a 300-candidate DINO-WM run, keeping linear weights packed between calls reduces persistent model state from 168.72 to 27.55 MB, matching Table~\ref{tab:quantization_resource}, while packed and expanded layouts produce identical candidate pools, elite sets, and final actions.

\section{Robustness of Cross-Model Transfer}
\label{app:transfer_controls}
Table~\ref{tab:cross_model_distortion} shows that direct W4 success does not rank sensitivity to transferred residuals. We test whether this reversal depends on the projection basis, on the different amplitudes of the two donors, or on the quantizer used to generate the LeWM residuals. All conditions follow Section~\ref{app:transfer_implementation}, perturb only the encoder interface of a fixed FP receiver, retain the receiver's native planner, and are evaluated on the 300 expert tasks.

\subsection{Random projections}
The cross-model maps in Table~\ref{tab:cross_model_distortion} use fixed random projections drawn from one seed. We redraw them with two additional seeds, keeping the donor residual banks and the amplitude rule of Equation~\eqref{eq:stransfer} unchanged. Self-transfer uses the identity map and is not repeated. Model and planner seeds are held fixed, so the three conditions differ only in the projection.

\begin{table}[!tbp]
\centering\small
\caption{Encoder-residual transfer under three random projections. Success (\%) over the 300 expert tasks; projection~1 is the one used in Table~\ref{tab:cross_model_distortion}.}
\label{tab:sseeds}
\setlength{\tabcolsep}{3.5pt}
\begin{tabular}{@{}ccccc@{}}
\toprule
Donor $\rightarrow$ receiver & Receiver FP & Proj.~1 & Proj.~2 & Proj.~3 \\
\midrule
LeWM $\rightarrow$ C-JEPA   & 85.0 & 83.3 & 85.0 & 86.3 \\
C-JEPA $\rightarrow$ LeWM   & 85.3 & 30.7 & 26.7 & 35.7 \\
C-JEPA $\rightarrow$ DDP-WM & 88.7 & 87.0 & 85.3 & 85.0 \\
\bottomrule
\end{tabular}
\end{table}

All three projections reproduce the pattern of Table~\ref{tab:cross_model_distortion} (Table~\ref{tab:sseeds}). C-JEPA residuals reduce LeWM to 26.7--35.7\%, far below both its FP success of 85.3\% and its direct-W4 success of 82.3\%, whereas DDP-WM retains 85.0--87.0\% under the same donor. LeWM residuals leave C-JEPA at 83.3--86.3\%, against an FP baseline of 85.0\%. The largest variation across projections, 9.0 points for C-JEPA$\rightarrow$LeWM, is small relative to the nearly 50-point gap between that receiver's FP and transferred success.

\subsection{Matched relative amplitude}
The two donors differ in amplitude as well as structure, with relative RMS values of approximately 1.07 for C-JEPA and 0.082 for LeWM. We therefore rescale each perturbation to a common relative RMS, measured against the receiver's mean-centered training-side FP latent RMS as in Equation~\eqref{eq:stransfer}. Scales are fixed on the training residual banks before evaluation, separately for current and goal representations. As a structure-free reference, we include an isotropic Gaussian perturbation at the same relative RMS, drawn from a fixed bank of 600 vectors per role with a random stream independent of the model and planner. Cross-model conditions use projection~1. We test a relative RMS of 1.0, close to the C-JEPA donor level, and 0.0818, the LeWM donor level.

\begin{table}[!tbp]
\centering\small
\caption{Encoder perturbations rescaled to a common relative RMS. Success (\%) over the 300 expert tasks. FP entries repeat Table~\ref{tab:cross_model_distortion}; dashes denote conditions not evaluated.}
\label{tab:smatched_rms}
\setlength{\tabcolsep}{4pt}
\begin{tabular}{@{}ccccc@{}}
\toprule
& & \multicolumn{3}{c}{Perturbation source} \\
\cmidrule(l){3-5}
Receiver & FP & C-JEPA & LeWM & Gaussian \\
\midrule
\multicolumn{5}{@{}l}{\emph{Relative RMS 1.0}} \\
C-JEPA & 85.0 & 53.3 & 62.0 & 59.0 \\
LeWM   & 85.3 & 30.7 & 28.3 & 28.3 \\
DDP-WM & 88.7 & 84.7 & 85.7 & 86.3 \\
\midrule
\multicolumn{5}{@{}l}{\emph{Relative RMS 0.0818}} \\
C-JEPA & 85.0 & 83.0 & 83.3 & 85.7 \\
LeWM   & 85.3 & 86.0 & 85.0 & 86.7 \\
DDP-WM & 88.7 & -- & -- & 88.7 \\
\bottomrule
\end{tabular}
\end{table}

At a common relative RMS of 1.0, the receiver accounts for most of the variation in planning success (Table~\ref{tab:smatched_rms}): for each source, success differs by 54.0--58.0 points across receivers, whereas for each receiver the three sources differ by at most 8.7 points. In particular, LeWM residuals rescaled to this amplitude reduce LeWM to 28.3\%, comparable to its 30.7\% under C-JEPA residuals. Since LeWM is the most sensitive receiver at this amplitude yet loses only 3.0 points under direct W4, its direct-W4 robustness stems from the small distortions its quantization produces rather than from tolerance to latent perturbations. Source dependence is largest for C-JEPA (53.3--62.0\%), the model for which Table~\ref{tab:smatched} also shows an effect of residual structure; the two comparisons match amplitude differently, by latent RMS here and by prediction error there.

At a relative RMS of 0.0818, every perturbation, including the isotropic Gaussian, leaves the receivers within 2.0 points of FP. The near-FP results under LeWM residuals in Table~\ref{tab:cross_model_distortion} therefore reflect their small amplitude. The same holds when LeWM residuals are generated with the native per-output-channel RTN of its direct-W4 result instead of G64 RTN: their relative RMS is 0.099, about one tenth of the C-JEPA donor level, and they leave LeWM and C-JEPA at 84.3\% and 85.0\%, within 1.0 point of FP.

Together, these controls show that the robustness reversal is stable across the tested random projections, persists when donor amplitude is matched, and does not depend on the quantizer used to generate the LeWM residuals. At a fixed amplitude, changing the receiver alters success far more than changing the perturbation source, supporting the separation of distortion generation and planning tolerance in Section~\ref{sec:cross_model_distortion}.